\pdfoutput=1

\documentclass[11pt]{article}

\usepackage[]{acl}
\usepackage{graphicx}
\usepackage{amssymb}
\usepackage{float}
\usepackage{tabularx, booktabs}
\usepackage{colortbl} 
\usepackage{multirow} 
\usepackage{multicol}
\usepackage{times}
\usepackage{latexsym}
\usepackage{graphicx}
\usepackage{listings}

\usepackage{color}
\usepackage{tcolorbox}
\usepackage{soul} 
\usepackage{CJKutf8}
\tcbuselibrary{listingsutf8} 
\usepackage{adjustbox}

\tcbset{width=0.9\textwidth,boxrule=0pt,colback=red,arc=0pt,auto outer arc,left=0pt,right=0pt,boxsep=5pt}

\usepackage{times}
\usepackage{latexsym}
\usepackage[T1]{fontenc}

\usepackage[utf8]{inputenc}

\usepackage{microtype}

\usepackage{inconsolata}

\usepackage{ulem}
\usepackage{xspace}
\usepackage{enumitem}
\usepackage{amsmath}
\usepackage{cleveref}
\usepackage{CJK}
\usepackage{hyperref}

\definecolor{TT}{HTML}{c7ceea}
\definecolor{VT}{HTML}{F9D0DA}
\definecolor{VV}{HTML}{c0dad1}
\definecolor{correct}{HTML}{007000}
\definecolor{wrong}{HTML}{cc0000}

\newcommand{\dataname}{\textit{\textbf{Chumor}}}

\newcommand{\chenv}[1]{\begin{CJK}{UTF8}{gbsn}#1\end{CJK}}
\usepackage{color}
\usepackage{mdframed}
\usepackage{tcolorbox}
\definecolor{lightgray}{rgb}{0.9,0.9,0.9}

\newtcolorbox{example}{
   boxrule=0pt,
   colback=white,
   colframe=white,
   boxsep=5pt,
   left=2pt,
   right=2pt,
   top=2pt,
   bottom=4pt,
   fontupper=\pcr\small,
   width=\linewidth,  
   before skip=0pt,   
   after skip=0pt,    
}

\title{\dataname~1.0: A Truly Funny and Challenging Chinese Humor Understanding Dataset from Ruo Zhi Ba}

\author{
    Ruiqi He$^1$ \quad 
    Yushu He$^1$ \quad
    Longju Bai$^{1}$ \quad
    {\bf Jiarui Liu$^{2}$ \quad}
    {\bf Zhenjie Sun$^{1}$ \quad}\\
    {\bf Zenghao Tang$^{3}$ \quad}
    {\bf He Wang$^{2}$ \quad}
    {\bf Hanchen Xia$^{3}$ \quad}
    {\bf Naihao Deng$^{1\star}$ \quad}\\
    $^{1}$University of Michigan\quad
    $^{2}$Carnegie Mellon University\quad
    $^{3}$Shanghai Jiaotong University \\
    {\tt \{ruiqih, dnaihao\}@umich.edu}
}

\begin{document}
\maketitle

\def\thefootnote{$\star$}\footnotetext{Corresponding author of this work.}

\begin{abstract}
    Existing humor datasets and evaluations predominantly focus on English, lacking resources for culturally nuanced humor in non-English languages like Chinese. 
    To address this gap, we construct \dataname, a dataset sourced from Ruo Zhi Ba (RZB, \chenv{弱智吧}), a Chinese Reddit-like platform dedicated to sharing intellectually challenging and culturally specific jokes. 
    We annotate explanations for each joke and evaluate human explanations against two state-of-the-art LLMs, GPT-4o and ERNIE Bot, through A/B testing by native Chinese speakers. 
    Our evaluation shows that \dataname~is challenging even for SOTA LLMs, and the human explanations for \dataname~jokes are significantly better than explanations generated by the LLMs. 
    We release \dataname~at \url{https://github.com/dnaihao/Chumor-dataset}.
\end{abstract}

\section{Introduction}


Humor is an intrinsic human trait that touches the core of our social and emotional lives, making it a rich field of study across various disciplines \cite{lefcourt2001humor, mihalcea-strapparava-2005-making, gelkopf2011use, hessel-etal-2023-androids}. 
With the advent of Large Language Models (LLMs), researchers have evaluated LLMs' performance on diverse tasks \cite{liu2023evaluating, deng2024tables, wu-etal-2023-hi}
and observed LLMs' extraordinary performance on many \cite{10.1162/tacl_a_00632}.
In contrast, researchers have observed that LLMs still fail to understand humor \cite{ghanadian-etal-2023-chatgpt}.
However, with all these studies on humor and LLMs' understanding of humor, most of these humor datasets and evaluations remain in English \cite{radev-etal-2016-humor, hasan-etal-2019-ur}.
This presents a significant gap, particularly for non-English languages like Chinese, where culturally nuanced humor understanding is underrepresented. 

\begin{figure}[t]
    \centering
    \includegraphics[width=\linewidth]{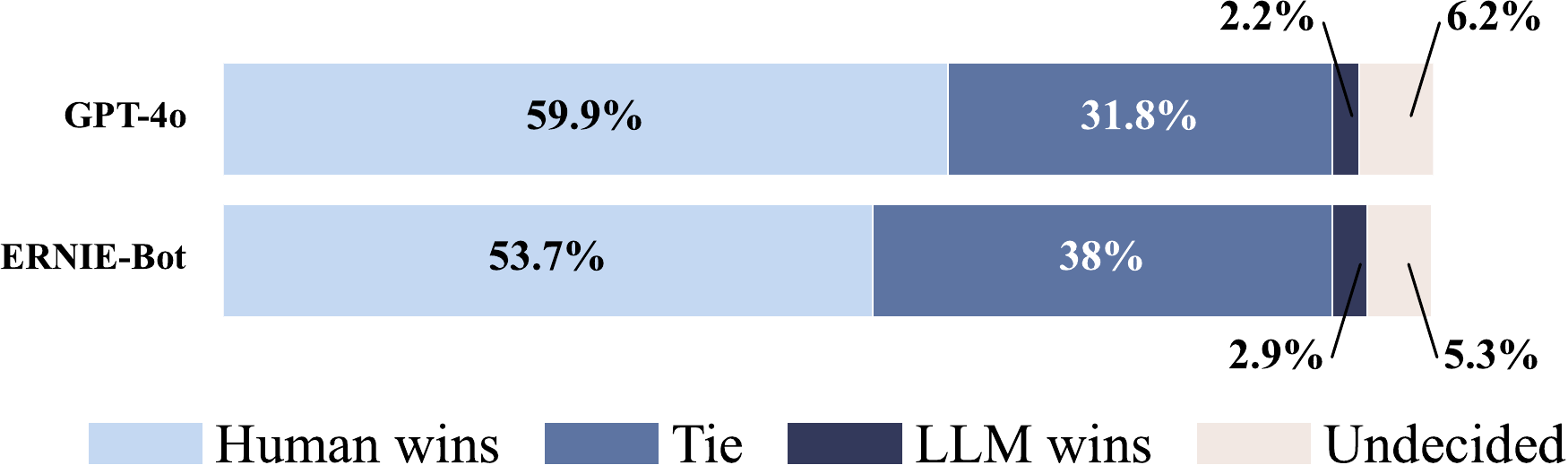}
    \caption{Annotated preference for whether human explanation is better (``Human wins'') or the explanation from LLMs is better (``LLM wins'').
    }
    \label{fig:gpt-4o-preference-eval}
\end{figure}

\begin{figure*}
    \centering
    \includegraphics[width=0.98\textwidth]{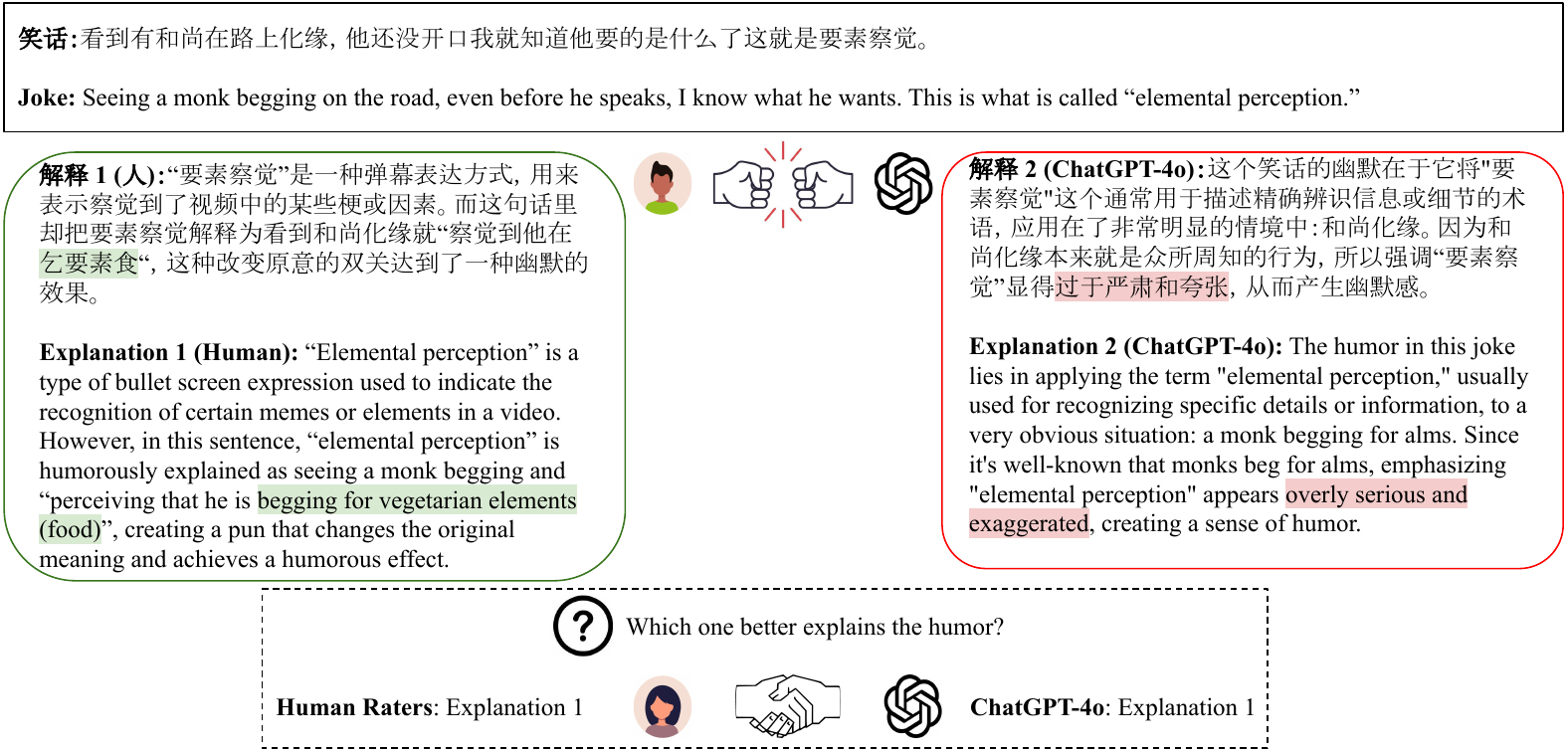}
    \caption{An Example of the Chinese joke from RZB (\begin{CJK}{UTF8}{gbsn}RZB, ``弱智吧''\end{CJK}) where the explanation differs between humans and ChatGPT-4o (as of June 5th, 2024).
    Interestingly, when we provide both explanations to ChatGPT-4o, it decides the human explanation better explains the humor than its own explanation.
    This agrees with the choice of human raters who also decide the explanation from human better explains the humor.
    We include a further discussion of whether LLMs can serve as the preference annotator in \Cref{app-sec: llms-as-preference-annotator}.}
    \label{fig:example}
\end{figure*}

In this paper, we try to address this gap by constructing \dataname, a truly funny and challenging Chinese humor understanding dataset sourced from Ruo Zhi Ba (\begin{CJK}{UTF8}{gbsn}RZB, ``弱智吧''\end{CJK} in Chinese), a Chinese version of Reddit platform dedicated to sharing intellectually challenging and culturally specific jokes.
This platform provides a set of unique Chinese jokes that incorporate the subtleties and intricacies of Chinese humor.
\Cref{fig:example} provides an example of the joke from RZB.
In addition, \citet{bai2024coig} have observed that when tuning LLMs on data from RZB, LLMs achieve the best performance on Chinese reasoning tasks compared to tuning LLMs on data from other sources, indicating the significant value of jokes from RZB.

In \dataname, we manually annotate the explanations for each joke.
We then prompt two state-of-the-art (SOTA) LLMs, GPT-4o from OpenAI and ERNIE Bot from Baidu to get their explanations.
In the evaluation, a group of native Chinese speakers determine their preferences between the explanations from human and the LLM in an A/B testing fashion.
We note that in the preference annotation process, all of our annotators report to us that the jokes are very funny and align well with the Chinese Internet trends.
Our evaluation shows that \dataname~is challenging even for the SOTA LLMs, and the human explanations for \dataname~jokes are significantly better than explanations generated by the LLMs (\Cref{fig:gpt-4o-preference-eval}).
In addition, we provide examples of our jokes on which LLMs fail to provide explanations and the hypothesis of their failures.

In summary, our contributions are two folds:

\begin{enumerate}[leftmargin=\parindent,align=left,labelwidth=\parindent,labelsep=0pt]
    \item We construct \dataname, a truly funny and challenging Chinese humor understanding dataset, addressing the lack of non-English humor understanding dataset.
    \item We reveal that on \dataname, human explanations for jokes are significantly better than explanations from SOTA LLMs.
\end{enumerate}

\section{Dataset Construction}

\paragraph{Data Collection.}
We construct \dataname~by including RZB jokes from ``Best Annual Threads'' between 2018 and 2021 that have been crawled previously\footnote{https://github.com/Leymore/ruozhiba}, and we also collect all threads in the ``Moderator's Recommendation'' section from RZB directly.
Each thread in RZB consists of \begin{CJK}{UTF8}{gbsn}``标题'' (the title)\end{CJK}, \begin{CJK}{UTF8}{gbsn}``一楼'' (the content)\end{CJK}, and several "\begin{CJK}{UTF8}{gbsn}``跟帖'' (the follow-up posts)\end{CJK}. 
For threads from Best Annual Threads, the jokes are listed in the follow-up posts, which are selected by the forum moderator.
For threads from Moderator's Recommendation, the jokes consist of the title and the content of each thread.
We remove the content if it repeats the title.

\paragraph{Data Cleaning.}
We store both the title and the content of the raw data.
However, because the posting restrictions of the platform require that the content cannot be empty, many posts contain meaningless placeholder texts such as ``.'', ``!'', ``0'', ``RT'', among others. 
We automatically identify and remove these patterns, and only keep the title which is the joke itself.
In the meantime, due to the length limitations on the original platform, many post titles are truncated from the beginning parts of the content. 
We identify these instances and replace the truncated title with the complete content to get the actual joke.
We also remove duplicates that appear both in the 
``Moderator's Recommendation'' and the ``Best Annual Posts''.

\paragraph{Data Annotation.}
Since the crowd-sourcing typically cannot solicit high-quality explanations according to \citet{hessel-etal-2023-androids}, one of the authors decides to annotate all the explanations to ensure the quality and consistency following \citet{hessel-etal-2023-androids}.
During the process, the author manually removes the threads that are not funny, threads related to forum management and rules, threads that include excessively offensive content, threads with incomplete content, and threads that focus more on philosophical insight rather than humor.

We note that this is a substantial effort, the author ended up annotating the explanations for 1,951 jokes and the resulting corpus has a mean of 78 Chinese characters of explanation per joke, and the total length, 151,730 Chinese characters, is comparable in length to a novella\footnote{The total length of our explanations surpasses the Chinese version of {\it The Great Gatsby} (100k Chinese characters), and is about half the length of the Chinese version of {\it Wuthering Heights} (325k Chinese characters).}.
\Cref{tab:he-dataset_comparison} provides a comparison between \dataname~and the existing humor explanation datasets, and to the best of our knowledge, \dataname~is the first Chinese humor explanation dataset.
We include a more comprehensive overview of the datasets related to humor in \Cref{tab:dataset_comparison} in \Cref{app-sec: dataset-comparison}. 




\begin{table}
  \small
  \centering
    \begin{tabular}{p{0.325\linewidth}p{0.325\linewidth}cc}
      \toprule
      Dataset & Source & L & \# J\\
      \midrule
      ExPUN \cite{sun-etal-2022-expunations} &  SemEval Task 7  & en & 1,999 \\
      Joke explanation \footnotemark  &  explainthejoke.com  & en & 377 \\
      NYT-Captions \cite{hessel-etal-2023-androids} & New Yorker caption contest & en & 651 \\
      \dataname~(us) & {\bf Ruo Zhi Ba} & {\bf zh} & {\bf 1,951}\\
      \bottomrule
    \end{tabular}
  \caption{Comparison between our collected \dataname~dataset and other humor explanation datasets.
  ``L'' represents the language of the dataset (en: English, zh: Chinese).
  ``\# J'' represents the number of manually annotated explanations in each dataset.}
  \label{tab:he-dataset_comparison}
\end{table}

\footnotetext{https://huggingface.co/violetamaral/joke-explaination/}

\section{Experiments}

\paragraph{Experiment Setup.}
We select two SOTA LLMs, one LLM from the company in western world, GPT-4o\footnote{https://openai.com/index/hello-gpt-4o/} from OpenAI, and the other from a Chinese Internet company, ERNIE Bot\footnote{http://research.baidu.com/Blog/index-view?id=174} from Baidu.
To evaluate the innate Chinese humor understanding abilities of both LLMs, We prompt them in a zero-shot setting to explain the humor in two sentences as all the human explanations are in two sentences.
Here is the prompt we feed to both LLMs: \chenv{请用两句话解释这个笑话的幽默之处:\textbackslash n[Joke]}, which translates to ``Please explain the joke in two sentences:\textbackslash n[Joke]''.

\paragraph{Evaluation Setup.}
To fairly evaluate which explanation is better, we conduct an A/B testing by presenting the humor explanation from one LLM and from human to six college students, and ask them to annotate their preference of the explanation for each joke.
These college students are native Chinese speakers who grow up in China, therefore they have a deep understanding of the cultural terms and trending terms in China.
We note that the preference annotation requires a substantial effort as each annotator reads through a total length of around 300k Chinese characters\footnote{This is about the same length of the Chinese version of {\it Wuthering Heights} (325k Chinese characters).}.
We end up with three preference annotations for each joke.
For the preference annotations, we achieve a 61.39\% agreement rate among annotators (\Cref{app-sec: agreement-rate}).

We employ the winning rate as our measure to compare LLMs' explanation versus human explanation.
We take the majority vote among all annotators for each example.
In addition, if all annotators disagree, we assign an ``Undecided'' label.
\Cref{app-sec: annotation-instruction} provides the annotation instructions we present to the annotators.

\paragraph{Overall Results.}

\begin{figure}[t]
    \centering
    \includegraphics[width=\linewidth]{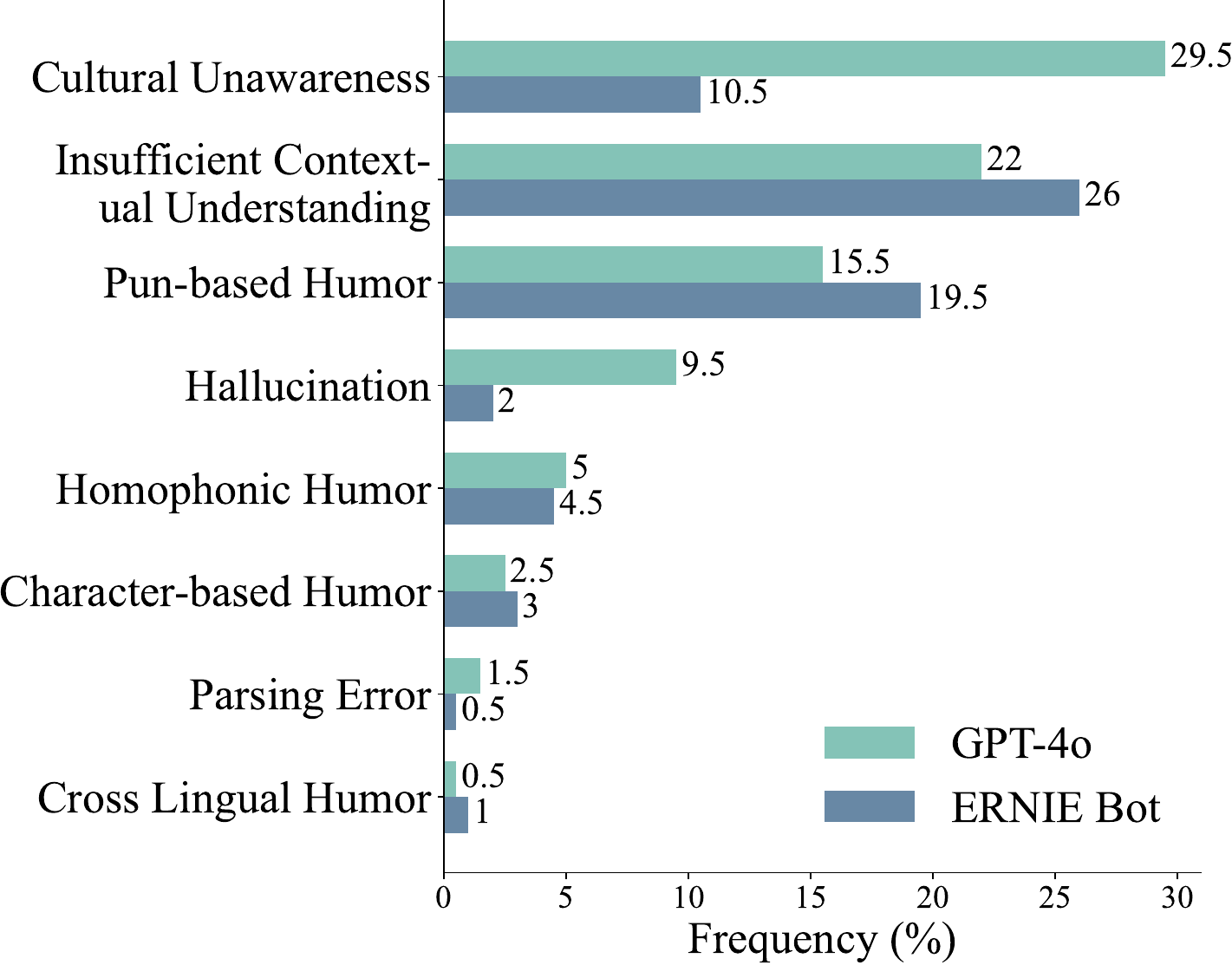}
    \caption{Distribution of error types for GPT-4o and ERNIE Bot. 
    We sample 200 examples to calculate the distribution of these error types. 
    We note that an example may correspond to multiple error types. 
    }
    \label{fig:distribution}
\end{figure}

\Cref{fig:gpt-4o-preference-eval} reports the wining rate of explanations from human versus GPT-4o and ERNIE Bot.
We can see that human explanations are significantly better than explanations from both LLMs, with human winning over 50\% of the time, and LLMs win in 2-3\% of cases.



\section{Error Analysis}

\Cref{fig:distribution} provides an overall distribution of error types for GPT-4o and ERNIE Bot on \dataname~in terms of their humor explanations.
GPT-4o exhibits a significantly higher error rate on jokes related to Chinese culture (29.5\% compared to 10.5\% for ERNIE Bot on cultural unawareness). 
We suspect that ERNIE Bot is more familiar with Chinese culture as it may be trained with a larger Chinese corpus than GPT-4o.
GPT-4o performs better on cases that require an understanding of the context or the puns, suggesting its strong reasoning ability.
We provide three error cases for GPT-4o here and more cases for GPT-4o and ERNIE Bot in \Cref{app-sec: more-error-cases}. 
In the following examples, we underscore the key phrases that induce humor in green, and underscore the errors in red. 

\paragraph{Cultural Unawareness.}\leavevmode

\noindent\includegraphics[width=\linewidth]{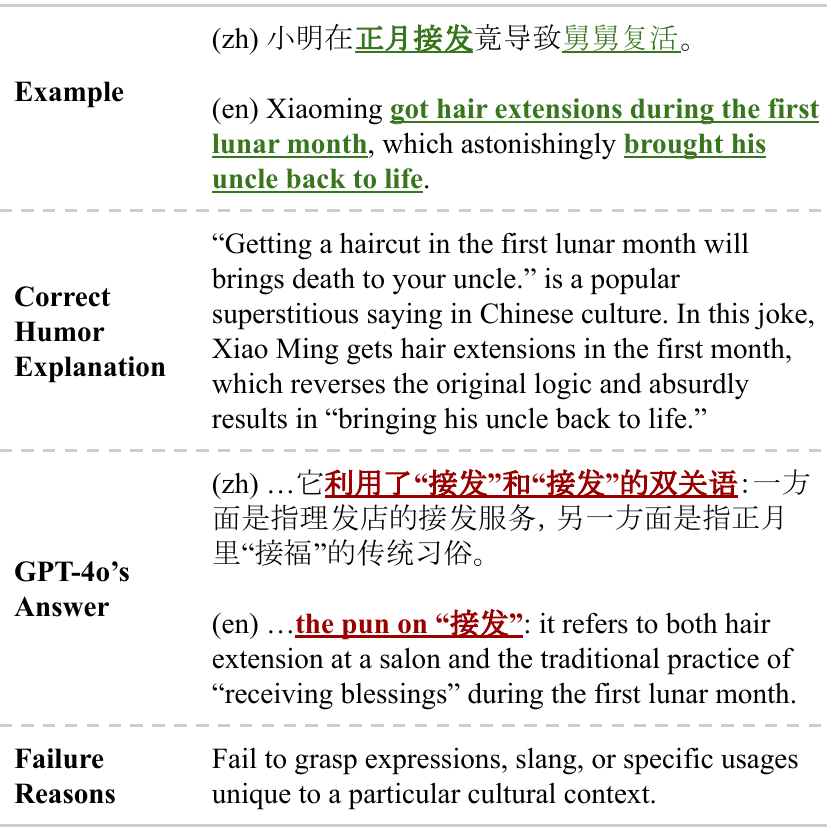}
\noindent LLMs may fail to explain the joke because they are not aware of certain cultural knowledge.
For instance, this example requires the knowledge of a superstitious belief in Chinese culture, {\it getting a haircut in the first lunar month brings death to your uncle}, and the explanation from GPT-4o fails to connect to this Chinese cultural belief.
Though LLMs have been pre-trained on Internet-scale corpus, such culturally specific knowledge can still be very challenging for them to grasp.
Moreover, even if they have acquired such cultural knowledge, they may fail to relate to them as we humans do in the reasoning process.

\paragraph{Homophonic Humor.}\leavevmode

\noindent\includegraphics[width=\linewidth]{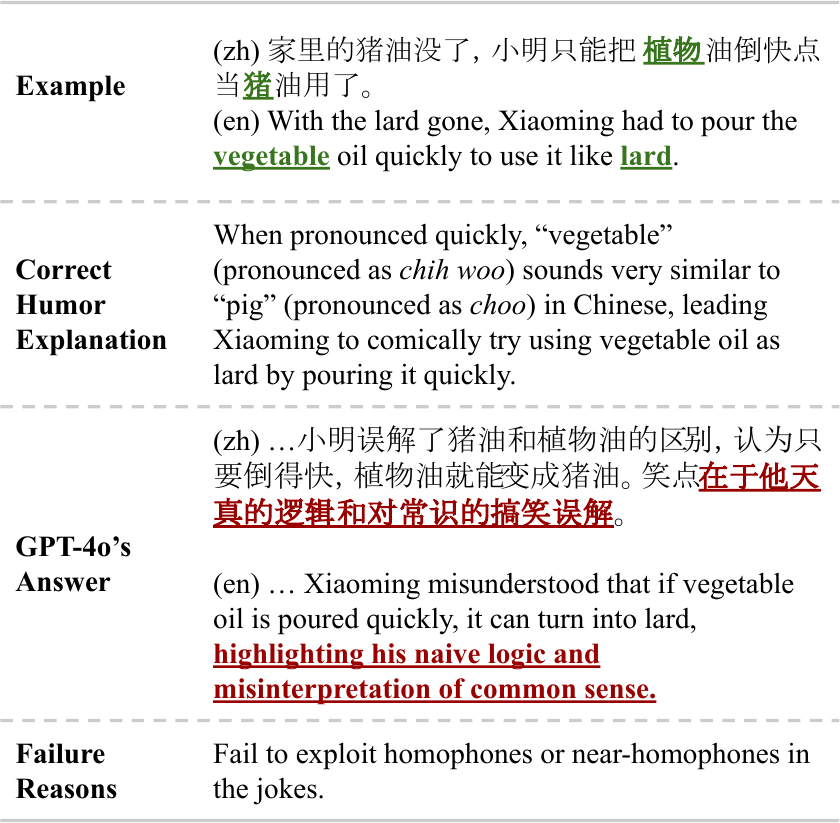}

\noindent LLMs may fail to exploit homophones or near-homophones in the joke.
This example requires LLMs to reason over the pronunciation as ``\chenv{植物}'' (pronounced as {\it chih woo}, meaning ``vegetable'') sounds very similar to ``\chenv{猪}'' (pronounced as {\it choo}, meaning ``pig'') in Chinese when we speak it fast enough, and the humor arises from the contrast between how similarly the items are pronounced and how unrelated they actually are.
Such contrast may be sparse in the training corpus for LLMs, and also requires deep connections across different modalities to make LLMs understand the linkage between pronunciation and the meaning behind these terms.

\paragraph{Character-based Humor.}\leavevmode

\noindent\includegraphics[width=\linewidth]{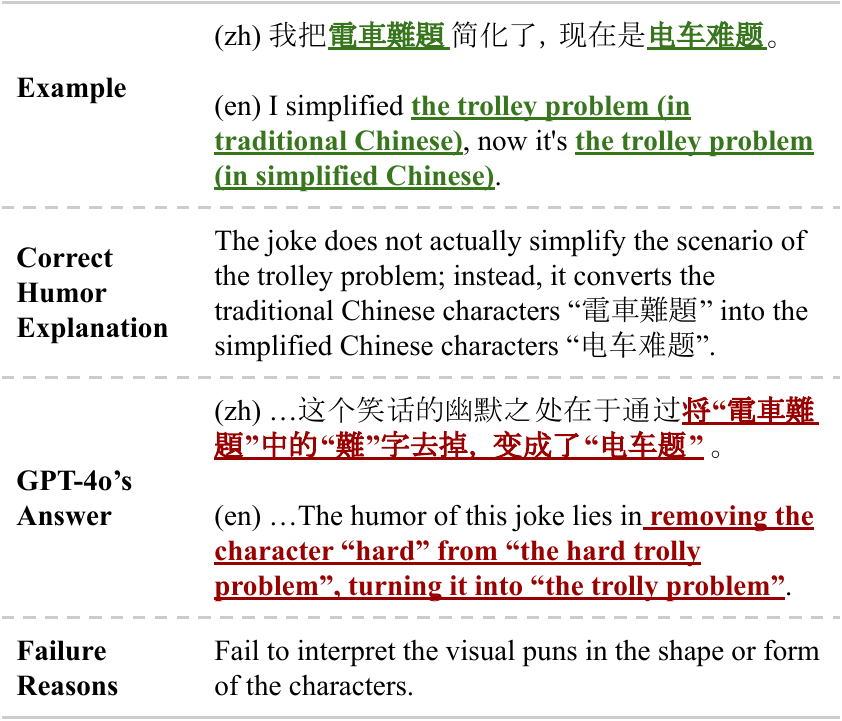}

\noindent LLMs may fail to interpret the visual puns in the shape or form for Chinese characters.
In this example, ``simplify'' does not mean to simplify the trolly problem, but to simplify the traditional Chinese characters to simplified Chinese characters as the traditional Chinese characters are also termed as ``complicated characters''.
However, it is difficult for LLMs to reason such graphemic differences as there are no explicit connections between the textual meaning and visual representations of the glyphs.

\section{Conclusion}

We present \dataname, a Chinese humor understanding dataset that includes intellectually challenging and culturally specific humor in Chinese.
We have shown that \dataname~is challenging even for the two state-of-the-art LLMs and provided analysis of their failure cases.
We hope that \dataname~can facilitate non-English humor research and the research on enhancing LLMs' reasoning abilities for diverse cultural backgrounds.

\section*{Limitations}
Due to the time consumption and availability of native Chinese speakers, we did not include a large-scale preference annotation.
We encourage future research on collecting large-scale preference data, especially for non-English languages. 
Moreover, humor is highly subjective, therefore a large-scale preference annotation may also benefit the models \cite{deng-etal-2023-annotate}.
However, we want to stress the high quality of our preference data annotations, as all of our annotators are native Chinese speakers and hold at least a bachelor's degree or are involved in bachelor's programs.
We also encourage future research on comprehensive evaluation of LLMs' humor understanding abilities, especially open-source LLMs' humor understanding abilities.
In the meantime, we want to stress that our research focuses primarily on revealing how humor understanding remains challenging even for the SOTA LLMs. 
Our work shows that along with many other problems \cite{ignat-etal-2024-solved-open}, humor understanding, especially non-English and culturally specific humor understanding, remains an unsolved problem in the LLM era. 
We hope \dataname~can contribute to non-English humor understanding evaluations for future multi-lingual LLMs.

\section*{Ethics Statement}
We tried our best to filter out the excessively offensive content in RZB.
However, due to the subjective nature of humor, some of our jokes may be considered offensive to certain people due to different standards.
Therefore, we recommend researchers to use \dataname~with cultural sensitivity and an understanding that the jokes reflect the sociocultural context in which they were created.
We encourage users of \dataname \ to maintain caution when working with the dataset, and to be mindful of the potential for offense or harm, especially when using the dataset for research or applications involving diverse audiences or sensitive domains.
We provide some potential usages of \dataname~in \Cref{app-sec: potential-application}.
We wish to foster an ethical and responsible approach to data collection and usage, and we welcome constructive feedback from the research community and stakeholders to continually improve \dataname~and mitigate potential harm.

\section{Acknowledgement}
The GPT-4o experiments are supported by the credit from OpenAI through OpenAI Researcher Access. 
We appreciate the review and feedback from Sheng Zhang.


\bibliography{custom, aclanthology}

\begin{thebibliography}{41}
\expandafter\ifx\csname natexlab\endcsname\relax\def\natexlab#1{#1}\fi

\bibitem[{Bai et~al.(2024)Bai, Du, Liang, Jin, Liu, Zhou, Zheng, Zhang, Ma, Wang et~al.}]{bai2024coig}
Yuelin Bai, Xinrun Du, Yiming Liang, Yonggang Jin, Ziqiang Liu, Junting Zhou, Tianyu Zheng, Xincheng Zhang, Nuo Ma, Zekun Wang, et~al. 2024.
\newblock Coig-cqia: Quality is all you need for chinese instruction fine-tuning.
\newblock \emph{arXiv preprint arXiv:2403.18058}.

\bibitem[{Bertero and Fung(2016)}]{bertero-fung-2016-deep}
Dario Bertero and Pascale Fung. 2016.
\newblock \href {https://aclanthology.org/L16-1079} {Deep learning of audio and language features for humor prediction}.
\newblock In \emph{Proceedings of the Tenth International Conference on Language Resources and Evaluation ({LREC}'16)}, pages 496--501, Portoro{\v{z}}, Slovenia. European Language Resources Association (ELRA).

\bibitem[{Castro et~al.(2019)Castro, Hazarika, P{\'e}rez-Rosas, Zimmermann, Mihalcea, and Poria}]{castro-etal-2019-towards}
Santiago Castro, Devamanyu Hazarika, Ver{\'o}nica P{\'e}rez-Rosas, Roger Zimmermann, Rada Mihalcea, and Soujanya Poria. 2019.
\newblock \href {https://doi.org/10.18653/v1/P19-1455} {Towards multimodal sarcasm detection (an {\_}{O}bviously{\_} perfect paper)}.
\newblock In \emph{Proceedings of the 57th Annual Meeting of the Association for Computational Linguistics}, pages 4619--4629, Florence, Italy. Association for Computational Linguistics.

\bibitem[{Chen and Lee(2017)}]{chen-lee-2017-predicting}
Lei Chen and Chong~Min Lee. 2017.
\newblock \href {https://doi.org/10.18653/v1/W17-5009} {Predicting audience{'}s laughter during presentations using convolutional neural network}.
\newblock In \emph{Proceedings of the 12th Workshop on Innovative Use of {NLP} for Building Educational Applications}, pages 86--90, Copenhagen, Denmark. Association for Computational Linguistics.

\bibitem[{Chen et~al.(2023)Chen, Li, Liang, Xiao, Liu, and Chen}]{10.1145/3539597.3570431}
Yuyan Chen, Zhixu Li, Jiaqing Liang, Yanghua Xiao, Bang Liu, and Yunwen Chen. 2023.
\newblock \href {https://doi.org/10.1145/3539597.3570431} {Can pre-trained language models understand chinese humor?}
\newblock In \emph{Proceedings of the Sixteenth ACM International Conference on Web Search and Data Mining}, WSDM '23, page 465–480, New York, NY, USA. Association for Computing Machinery.

\bibitem[{Chen et~al.(2024)Chen, Yuan, Liu, Liu, Guan, Guo, Peng, Liu, Li, and Xiao}]{Chen_Yuan_Liu_Liu_Guan_Guo_Peng_Liu_Li_Xiao_2024}
Yuyan Chen, Yichen Yuan, Panjun Liu, Dayiheng Liu, Qinghao Guan, Mengfei Guo, Haiming Peng, Bang Liu, Zhixu Li, and Yanghua Xiao. 2024.
\newblock \href {https://doi.org/10.1609/aaai.v38i16.29736} {Talk funny! a large-scale humor response dataset with chain-of-humor interpretation}.
\newblock \emph{Proceedings of the AAAI Conference on Artificial Intelligence}, 38(16):17826--17834.

\bibitem[{Deng et~al.(2024)Deng, Sun, He, Sikka, Chen, Ma, Zhang, and Mihalcea}]{deng2024tables}
Naihao Deng, Zhenjie Sun, Ruiqi He, Aman Sikka, Yulong Chen, Lin Ma, Yue Zhang, and Rada Mihalcea. 2024.
\newblock Tables as images? exploring the strengths and limitations of llms on multimodal representations of tabular data.
\newblock \emph{arXiv preprint arXiv:2402.12424}.

\bibitem[{Deng et~al.(2023)Deng, Zhang, Liu, Wu, Wang, and Mihalcea}]{deng-etal-2023-annotate}
Naihao Deng, Xinliang Zhang, Siyang Liu, Winston Wu, Lu~Wang, and Rada Mihalcea. 2023.
\newblock \href {https://doi.org/10.18653/v1/2023.findings-emnlp.832} {You are what you annotate: Towards better models through annotator representations}.
\newblock In \emph{Findings of the Association for Computational Linguistics: EMNLP 2023}, pages 12475--12498, Singapore. Association for Computational Linguistics.

\bibitem[{Du et~al.(2024)Du, Yu, Gao, Pan, Cheng, Ma, Yuan, Qu, Liu, Zheng et~al.}]{du2024chinese}
Xinrun Du, Zhouliang Yu, Songyang Gao, Ding Pan, Yuyang Cheng, Ziyang Ma, Ruibin Yuan, Xingwei Qu, Jiaheng Liu, Tianyu Zheng, et~al. 2024.
\newblock Chinese tiny llm: Pretraining a chinese-centric large language model.
\newblock \emph{arXiv preprint arXiv:2404.04167}.

\bibitem[{Engelthaler and Hills(2018)}]{engelthaler2018humor}
Tomas Engelthaler and Thomas~T Hills. 2018.
\newblock Humor norms for 4,997 english words.
\newblock \emph{Behavior research methods}, 50:1116--1124.

\bibitem[{Fu et~al.(2023)Fu, Ng, Jiang, and Liu}]{fu2023gptscore}
Jinlan Fu, See-Kiong Ng, Zhengbao Jiang, and Pengfei Liu. 2023.
\newblock Gptscore: Evaluate as you desire.
\newblock \emph{arXiv preprint arXiv:2302.04166}.

\bibitem[{Gelkopf et~al.(2011)}]{gelkopf2011use}
Marc Gelkopf et~al. 2011.
\newblock The use of humor in serious mental illness: A review.
\newblock \emph{Evidence-Based Complementary and Alternative Medicine}, 2011.

\bibitem[{Ghanadian et~al.(2023)Ghanadian, Nejadgholi, and Al~Osman}]{ghanadian-etal-2023-chatgpt}
Hamideh Ghanadian, Isar Nejadgholi, and Hussein Al~Osman. 2023.
\newblock \href {https://doi.org/10.18653/v1/2023.wassa-1.16} {{C}hat{GPT} for suicide risk assessment on social media: Quantitative evaluation of model performance, potentials and limitations}.
\newblock In \emph{Proceedings of the 13th Workshop on Computational Approaches to Subjectivity, Sentiment, {\&} Social Media Analysis}, pages 172--183, Toronto, Canada. Association for Computational Linguistics.

\bibitem[{Hasan et~al.(2019)Hasan, Rahman, Bagher~Zadeh, Zhong, Tanveer, Morency, and Hoque}]{hasan-etal-2019-ur}
Md~Kamrul Hasan, Wasifur Rahman, AmirAli Bagher~Zadeh, Jianyuan Zhong, Md~Iftekhar Tanveer, Louis-Philippe Morency, and Mohammed~(Ehsan) Hoque. 2019.
\newblock \href {https://doi.org/10.18653/v1/D19-1211} {{UR}-{FUNNY}: A multimodal language dataset for understanding humor}.
\newblock In \emph{Proceedings of the 2019 Conference on Empirical Methods in Natural Language Processing and the 9th International Joint Conference on Natural Language Processing (EMNLP-IJCNLP)}, pages 2046--2056, Hong Kong, China. Association for Computational Linguistics.

\bibitem[{Hessel et~al.(2023)Hessel, Marasovic, Hwang, Lee, Da, Zellers, Mankoff, and Choi}]{hessel-etal-2023-androids}
Jack Hessel, Ana Marasovic, Jena~D. Hwang, Lillian Lee, Jeff Da, Rowan Zellers, Robert Mankoff, and Yejin Choi. 2023.
\newblock \href {https://doi.org/10.18653/v1/2023.acl-long.41} {Do androids laugh at electric sheep? humor {``}understanding{''} benchmarks from the new yorker caption contest}.
\newblock In \emph{Proceedings of the 61st Annual Meeting of the Association for Computational Linguistics (Volume 1: Long Papers)}, pages 688--714, Toronto, Canada. Association for Computational Linguistics.

\bibitem[{Hossain et~al.(2019)Hossain, Krumm, and Gamon}]{hossain-etal-2019-president}
Nabil Hossain, John Krumm, and Michael Gamon. 2019.
\newblock \href {https://doi.org/10.18653/v1/N19-1012} {{``}president vows to cut {\textless}taxes{\textgreater} hair{''}: Dataset and analysis of creative text editing for humorous headlines}.
\newblock In \emph{Proceedings of the 2019 Conference of the North {A}merican Chapter of the Association for Computational Linguistics: Human Language Technologies, Volume 1 (Long and Short Papers)}, pages 133--142, Minneapolis, Minnesota. Association for Computational Linguistics.

\bibitem[{Ignat et~al.(2024)Ignat, Jin, Abzaliev, Biester, Castro, Deng, Gao, Gunal, He, Kazemi, Khalifa, Koh, Lee, Liu, Min, Mori, Nwatu, Perez-Rosas, Shen, Wang, Wu, and Mihalcea}]{ignat-etal-2024-solved-open}
Oana Ignat, Zhijing Jin, Artem Abzaliev, Laura Biester, Santiago Castro, Naihao Deng, Xinyi Gao, Aylin~Ece Gunal, Jacky He, Ashkan Kazemi, Muhammad Khalifa, Namho Koh, Andrew Lee, Siyang Liu, Do~June Min, Shinka Mori, Joan~C. Nwatu, Veronica Perez-Rosas, Siqi Shen, Zekun Wang, Winston Wu, and Rada Mihalcea. 2024.
\newblock \href {https://aclanthology.org/2024.lrec-main.708} {Has it all been solved? open {NLP} research questions not solved by large language models}.
\newblock In \emph{Proceedings of the 2024 Joint International Conference on Computational Linguistics, Language Resources and Evaluation (LREC-COLING 2024)}, pages 8050--8094, Torino, Italia. ELRA and ICCL.

\bibitem[{Lefcourt(2001)}]{lefcourt2001humor}
Herbert~M Lefcourt. 2001.
\newblock \emph{Humor: The psychology of living buoyantly}.
\newblock Springer Science \& Business Media.

\bibitem[{Li et~al.(2022)Li, Lin, Yang, Xu, and Zhang}]{10.1007/978-3-031-17120-8_41}
Zefeng Li, Hongfei Lin, Liang Yang, Bo~Xu, and Shaowu Zhang. 2022.
\newblock Memeplate: A chinese multimodal dataset for humor understanding in meme templates.
\newblock In \emph{Natural Language Processing and Chinese Computing}, pages 527--538, Cham. Springer International Publishing.

\bibitem[{Liu et~al.(2023{\natexlab{a}})Liu, Ning, Teng, Liu, Zhou, and Zhang}]{liu2023evaluating}
Hanmeng Liu, Ruoxi Ning, Zhiyang Teng, Jian Liu, Qiji Zhou, and Yue Zhang. 2023{\natexlab{a}}.
\newblock Evaluating the logical reasoning ability of chatgpt and gpt-4.
\newblock \emph{arXiv preprint arXiv:2304.03439}.

\bibitem[{Liu et~al.(2023{\natexlab{b}})Liu, Deng, Sabour, Jia, Huang, and Mihalcea}]{liu-etal-2023-task}
Siyang Liu, Naihao Deng, Sahand Sabour, Yilin Jia, Minlie Huang, and Rada Mihalcea. 2023{\natexlab{b}}.
\newblock \href {https://doi.org/10.18653/v1/2023.emnlp-main.944} {Task-adaptive tokenization: Enhancing long-form text generation efficacy in mental health and beyond}.
\newblock In \emph{Proceedings of the 2023 Conference on Empirical Methods in Natural Language Processing}, pages 15264--15281, Singapore. Association for Computational Linguistics.

\bibitem[{Liu et~al.(2021)Liu, Yang, Liu, Zhang, Luo, Zhang, Zhang, and Su}]{liu-etal-2021-bridging}
Xin Liu, Baosong Yang, Dayiheng Liu, Haibo Zhang, Weihua Luo, Min Zhang, Haiying Zhang, and Jinsong Su. 2021.
\newblock \href {https://doi.org/10.18653/v1/2021.acl-long.468} {Bridging subword gaps in pretrain-finetune paradigm for natural language generation}.
\newblock In \emph{Proceedings of the 59th Annual Meeting of the Association for Computational Linguistics and the 11th International Joint Conference on Natural Language Processing (Volume 1: Long Papers)}, pages 6001--6011, Online. Association for Computational Linguistics.

\bibitem[{Liu et~al.(2023{\natexlab{c}})Liu, Iter, Xu, Wang, Xu, and Zhu}]{liu-etal-2023-g}
Yang Liu, Dan Iter, Yichong Xu, Shuohang Wang, Ruochen Xu, and Chenguang Zhu. 2023{\natexlab{c}}.
\newblock \href {https://doi.org/10.18653/v1/2023.emnlp-main.153} {{G}-eval: {NLG} evaluation using gpt-4 with better human alignment}.
\newblock In \emph{Proceedings of the 2023 Conference on Empirical Methods in Natural Language Processing}, pages 2511--2522, Singapore. Association for Computational Linguistics.

\bibitem[{Mihalcea and Strapparava(2005)}]{mihalcea-strapparava-2005-making}
Rada Mihalcea and Carlo Strapparava. 2005.
\newblock \href {https://aclanthology.org/H05-1067} {Making computers laugh: Investigations in automatic humor recognition}.
\newblock In \emph{Proceedings of Human Language Technology Conference and Conference on Empirical Methods in Natural Language Processing}, pages 531--538, Vancouver, British Columbia, Canada. Association for Computational Linguistics.

\bibitem[{Panickssery et~al.(2024)Panickssery, Bowman, and Feng}]{panickssery2024llm}
Arjun Panickssery, Samuel~R Bowman, and Shi Feng. 2024.
\newblock Llm evaluators recognize and favor their own generations.
\newblock \emph{arXiv preprint arXiv:2404.13076}.

\bibitem[{Potash et~al.(2017)Potash, Romanov, and Rumshisky}]{potash-etal-2017-semeval}
Peter Potash, Alexey Romanov, and Anna Rumshisky. 2017.
\newblock \href {https://doi.org/10.18653/v1/S17-2004} {{S}em{E}val-2017 task 6: {\#}{H}ashtag{W}ars: Learning a sense of humor}.
\newblock In \emph{Proceedings of the 11th International Workshop on Semantic Evaluation ({S}em{E}val-2017)}, pages 49--57, Vancouver, Canada. Association for Computational Linguistics.

\bibitem[{Radev et~al.(2016)Radev, Stent, Tetreault, Pappu, Iliakopoulou, Chanfreau, de~Juan, Vallmitjana, Jaimes, Jha, and Mankoff}]{radev-etal-2016-humor}
Dragomir Radev, Amanda Stent, Joel Tetreault, Aasish Pappu, Aikaterini Iliakopoulou, Agustin Chanfreau, Paloma de~Juan, Jordi Vallmitjana, Alejandro Jaimes, Rahul Jha, and Robert Mankoff. 2016.
\newblock \href {https://aclanthology.org/L16-1076} {Humor in collective discourse: Unsupervised funniness detection in the new yorker cartoon caption contest}.
\newblock In \emph{Proceedings of the Tenth International Conference on Language Resources and Evaluation ({LREC}'16)}, pages 475--479, Portoro{\v{z}}, Slovenia. European Language Resources Association (ELRA).

\bibitem[{Sharma et~al.(2020)Sharma, Bhageria, Scott, PYKL, Das, Chakraborty, Pulabaigari, and Gamb{\"a}ck}]{sharma-etal-2020-semeval}
Chhavi Sharma, Deepesh Bhageria, William Scott, Srinivas PYKL, Amitava Das, Tanmoy Chakraborty, Viswanath Pulabaigari, and Bj{\"o}rn Gamb{\"a}ck. 2020.
\newblock \href {https://doi.org/10.18653/v1/2020.semeval-1.99} {{S}em{E}val-2020 task 8: Memotion analysis- the visuo-lingual metaphor!}
\newblock In \emph{Proceedings of the Fourteenth Workshop on Semantic Evaluation}, pages 759--773, Barcelona (online). International Committee for Computational Linguistics.

\bibitem[{Sun and Jurafsky(2004)}]{sun-jurafsky-2004-shallow}
Honglin Sun and Daniel Jurafsky. 2004.
\newblock \href {https://aclanthology.org/N04-1032} {Shallow semantic parsing of {C}hinese}.
\newblock In \emph{Proceedings of the Human Language Technology Conference of the North {A}merican Chapter of the Association for Computational Linguistics: {HLT}-{NAACL} 2004}, pages 249--256, Boston, Massachusetts, USA. Association for Computational Linguistics.

\bibitem[{Sun et~al.(2022)Sun, Narayan-Chen, Oraby, Cervone, Chung, Huang, Liu, and Peng}]{sun-etal-2022-expunations}
Jiao Sun, Anjali Narayan-Chen, Shereen Oraby, Alessandra Cervone, Tagyoung Chung, Jing Huang, Yang Liu, and Nanyun Peng. 2022.
\newblock \href {https://doi.org/10.18653/v1/2022.emnlp-main.304} {{E}x{PUN}ations: Augmenting puns with keywords and explanations}.
\newblock In \emph{Proceedings of the 2022 Conference on Empirical Methods in Natural Language Processing}, pages 4590--4605, Abu Dhabi, United Arab Emirates. Association for Computational Linguistics.

\bibitem[{Sun et~al.(2009)Sun, Sui, Wang, and Wang}]{sun-etal-2009-chinese}
Weiwei Sun, Zhifang Sui, Meng Wang, and Xin Wang. 2009.
\newblock \href {https://aclanthology.org/D09-1153} {{C}hinese semantic role labeling with shallow parsing}.
\newblock In \emph{Proceedings of the 2009 Conference on Empirical Methods in Natural Language Processing}, pages 1475--1483, Singapore. Association for Computational Linguistics.

\bibitem[{Tseng et~al.(2020)Tseng, Wu, Chang, Chen, and Hsu}]{tseng-etal-2020-development}
Yuen-Hsien Tseng, Wun-Syuan Wu, Chia-Yueh Chang, Hsueh-Chih Chen, and Wei-Lun Hsu. 2020.
\newblock \href {https://aclanthology.org/2020.lrec-1.168} {Development and validation of a corpus for machine humor comprehension}.
\newblock In \emph{Proceedings of the Twelfth Language Resources and Evaluation Conference}, pages 1346--1352, Marseille, France. European Language Resources Association.

\bibitem[{Wang et~al.(2022)Wang, Wu, Liu, Li, Tiwari, and Xie}]{Wang2022CanLM}
Benyou Wang, Xiang Wu, Xiaokang Liu, Jianquan Li, Prayag Tiwari, and Qianqian Xie. 2022.
\newblock \href {https://api.semanticscholar.org/CorpusID:250264242} {Can language models make fun? a case study in chinese comical crosstalk}.
\newblock In \emph{Annual Meeting of the Association for Computational Linguistics}.

\bibitem[{Weller and Seppi(2020)}]{weller-seppi-2020-rjokes}
Orion Weller and Kevin Seppi. 2020.
\newblock \href {https://aclanthology.org/2020.lrec-1.753} {The r{J}okes dataset: a large scale humor collection}.
\newblock In \emph{Proceedings of the Twelfth Language Resources and Evaluation Conference}, pages 6136--6141, Marseille, France. European Language Resources Association.

\bibitem[{Wu et~al.(2021)Wu, Lin, Yang, and Xu}]{10.1007/978-3-030-88480-2_49}
Jiaming Wu, Hongfei Lin, Liang Yang, and Bo~Xu. 2021.
\newblock \href {https://doi.org/10.1007/978-3-030-88480-2_49} {Mumor: A multimodal dataset for humor detection in conversations}.
\newblock In \emph{Natural Language Processing and Chinese Computing: 10th CCF International Conference, NLPCC 2021, Qingdao, China, October 13–17, 2021, Proceedings, Part I}, page 619–627, Berlin, Heidelberg. Springer-Verlag.

\bibitem[{Wu et~al.(2023)Wu, He, Jia, Mihalcea, Chen, and Deng}]{wu-etal-2023-hi}
Yufan Wu, Yinghui He, Yilin Jia, Rada Mihalcea, Yulong Chen, and Naihao Deng. 2023.
\newblock \href {https://doi.org/10.18653/v1/2023.findings-emnlp.717} {Hi-{T}o{M}: A benchmark for evaluating higher-order theory of mind reasoning in large language models}.
\newblock In \emph{Findings of the Association for Computational Linguistics: EMNLP 2023}, pages 10691--10706, Singapore. Association for Computational Linguistics.

\bibitem[{Yang et~al.(2015)Yang, Lavie, Dyer, and Hovy}]{yang-etal-2015-humor}
Diyi Yang, Alon Lavie, Chris Dyer, and Eduard Hovy. 2015.
\newblock \href {https://doi.org/10.18653/v1/D15-1284} {Humor recognition and humor anchor extraction}.
\newblock In \emph{Proceedings of the 2015 Conference on Empirical Methods in Natural Language Processing}, pages 2367--2376, Lisbon, Portugal. Association for Computational Linguistics.

\bibitem[{Zhang et~al.(2019)Zhang, Zhang, Liu, Lin, and Xia}]{Zhang2019TellingTW}
Dongyu Zhang, Heting Zhang, Xikai Liu, Hongfei Lin, and Feng Xia. 2019.
\newblock \href {https://api.semanticscholar.org/CorpusID:202767593} {Telling the whole story: A manually annotated chinese dataset for the analysis of humor in jokes}.
\newblock In \emph{Conference on Empirical Methods in Natural Language Processing}.

\bibitem[{Zhang et~al.(2024{\natexlab{a}})Zhang, He, Ji, and Lu}]{zhang2024dont}
Min Zhang, Jianfeng He, Taoran Ji, and Chang-Tien Lu. 2024{\natexlab{a}}.
\newblock \href {http://arxiv.org/abs/2402.11406} {Don't go to extremes: Revealing the excessive sensitivity and calibration limitations of llms in implicit hate speech detection}.

\bibitem[{Zhang et~al.(2024{\natexlab{b}})Zhang, Ladhak, Durmus, Liang, McKeown, and Hashimoto}]{10.1162/tacl_a_00632}
Tianyi Zhang, Faisal Ladhak, Esin Durmus, Percy Liang, Kathleen McKeown, and Tatsunori~B. Hashimoto. 2024{\natexlab{b}}.
\newblock \href {https://doi.org/10.1162/tacl_a_00632} {{Benchmarking Large Language Models for News Summarization}}.
\newblock \emph{Transactions of the Association for Computational Linguistics}, 12:39--57.

\bibitem[{Zhao et~al.(2024)Zhao, Zhang, Zhang, Gui, and Huang}]{zhao2024llama}
Jun Zhao, Zhihao Zhang, Qi~Zhang, Tao Gui, and Xuanjing Huang. 2024.
\newblock Llama beyond english: An empirical study on language capability transfer.
\newblock \emph{arXiv preprint arXiv:2401.01055}.

\end{thebibliography}

\clearpage

\appendix

\section{Contributions}
\paragraph{Idea Proposal.}
Naihao Deng proposed the high-level idea of constructing a humor understanding benchmark sourced from RZB data.

\paragraph{Background Survey.} 
Ruiqi He surveyed the humor-related tasks. 

\paragraph{Data Processing.}
Ruiqi He crawled and processed the jokes from RZB.

\paragraph{Annotation.} 
Ruiqi He annotated the explanations for the RZB jokes.
Yushu He, Longju Bai, Jiarui Liu, Zhenjie Sun, Zhenghao Tang, He Wang, Naihao Deng conducted the preference annotations.

\paragraph{Experiments.}
Hanchen Xia conducted experiments to prompt ERNIE-Bot, Naihao Deng conducted experiments to prompt GPT-4o.

\paragraph{Result Aggregation.}
Ruiqi He, Naihao Deng, Yushu He aggregated the results.

\paragraph{Paper Writing.}
Ruiqi He and Naihao Deng drafted the paper.
Other authors provided revisions and feedback on the paper.

\noindent Naihao Deng organized the research.

\section{Dataset Comparison}
\label{app-sec: dataset-comparison}

\Cref{tab:dataset_comparison} provides a comprehensive overview of the existing datasets related to humor.
We note that \dataname~is the first humor explanation dataset in Chinese.

\begin{table*}
  \small
  \centering
    \begin{tabular}{p{0.27\linewidth}p{0.22\linewidth}cp{0.35\linewidth}}
      \toprule
      Dataset & Sources & Language & Tasks \\
      \midrule
      One Liners \cite{mihalcea-strapparava-2005-making} & Websites & en & Humor detection  \\
      Pun of the Day \cite{yang-etal-2015-humor} & Websites & en & Humor recognition, humor anchor extraction \\
      Big Bang Theory\cite{bertero-fung-2016-deep} & TV sitcom & en & Punchline detection\\
      Ted Laughter \cite{chen-lee-2017-predicting} & TED talks & en & Humor recognition, punchline detection \\
      \#HashtagWars \cite{potash-etal-2017-semeval} & TV show & en & Humor comparison and ranking \\
      HumorNorm \cite{engelthaler2018humor}& Online crowd-sourcing platform & en & Humor level rating \\
      MUStARD \cite{castro-etal-2019-towards}& Sitcoms & en & Sarcasm detection \\
      UR-FUNNY \cite{hasan-etal-2019-ur} & TED talks & en & Punchline detection \\
      Humicroedit \cite{hossain-etal-2019-president} & Reddit & en & Humor level rating and  generation \\
      rJokes \cite{weller-seppi-2020-rjokes} & Reddit & en & Sentiment analysis, humor level rating \\
      Memotion \cite{sharma-etal-2020-semeval}& Internet memes & en & Humor classification and level rating \\
      MUMOR \cite{10.1007/978-3-030-88480-2_49} &
      TV-sitcoms & en, zh & Humor detection \\
      NYT-Captions \cite{hessel-etal-2023-androids} & New Yorker caption contest & en & {\bf Humor explanation} \\
      Short Jokes Dataset\footnote{https://www.kaggle.com/datasets/abhinavmoudgil95/short-jokes} & Websites & en & - \\
      \midrule
      $C^{3}$ \cite{Wang2022CanLM} & Xiangsheng (Chinese crosstalk) & zh & Crosstalk generation \\
      TalkFunny \cite{Chen_Yuan_Liu_Liu_Guan_Guo_Peng_Liu_Li_Xiao_2024} & RED, Zhihu, etc & zh & Joke generation \\
      TCHD \cite{10.1145/3539597.3570431} & - & zh & Humor detection, level rating, classification, and punchline detection \\
      TTWS \cite{Zhang2019TellingTW} & Books, literary journals, etc & zh & Humorous word identification \\
      CHM \cite{tseng-etal-2020-development} & Websites, books, Apps & zh & Humor level rating and classification \\
      Memeplate \cite{10.1007/978-3-031-17120-8_41} & Social media, image recognition website & zh & Humor level rating \\
      \dataname~(us) & Ruo Zhi Ba & zh  & {\bf Humor explanation} \\
      \bottomrule
    \end{tabular}
  \caption{Existing datasets related to humor.
  We note that \dataname~is the first Chinese humor explanation dataset.}
\label{tab:dataset_comparison}
\end{table*}

\section{Agreement Rate Calculation}
\label{app-sec: agreement-rate}
We calculate the percentage agreement rate among annotators who annotate their preferences between explanations from LLMs and humans.
The results showed an average inter-annotator agreement of 61.92\% for GPT-4o and 60.86\% for ERNIE Bot. 
Considering the inherent subjectivity of humor interpretation tasks \cite{deng-etal-2023-annotate}, we consider that the combined average agreement percentage of 61.38\% is decent.

\section{Annotation Instructions for Preference Annotation}
\label{app-sec: annotation-instruction}

\noindent We include the following instructions for the preference annotations of the joke explanations:









\chenv{
\noindent ``在这个标注中，你将会看到一个笑话和对这个笑话的幽默之处的两个解释，请你比较哪个解释更好的解释了这个笑话的幽默之处，并从以下三个标签中选择：

\noindent 1. 解释1 

\noindent 2. 解释2 

\noindent 3. 一样好''}

\noindent which translates to,

\noindent ``In this annotation task, you will see a joke along with two explanations of its humor. 
Please compare which explanation better explains the reason why this joke is funny and choose from the following three labels:

\noindent 1. Explanation 1

\noindent 2. Explanation 2

\noindent 3. Both are equally good.''

For each example, we randomly assign the explanations from the LLM and the human as Explanation 1 and Explanation 2 to ensure a fair comparison.

\section{More Error Cases}
\label{app-sec: more-error-cases}

We note that many examples here encompass multiple error types, which demonstrates \dataname's complexity.

\paragraph{Insufficient Contextual Understanding.}\leavevmode
%
%

\noindent\includegraphics[width=\linewidth]{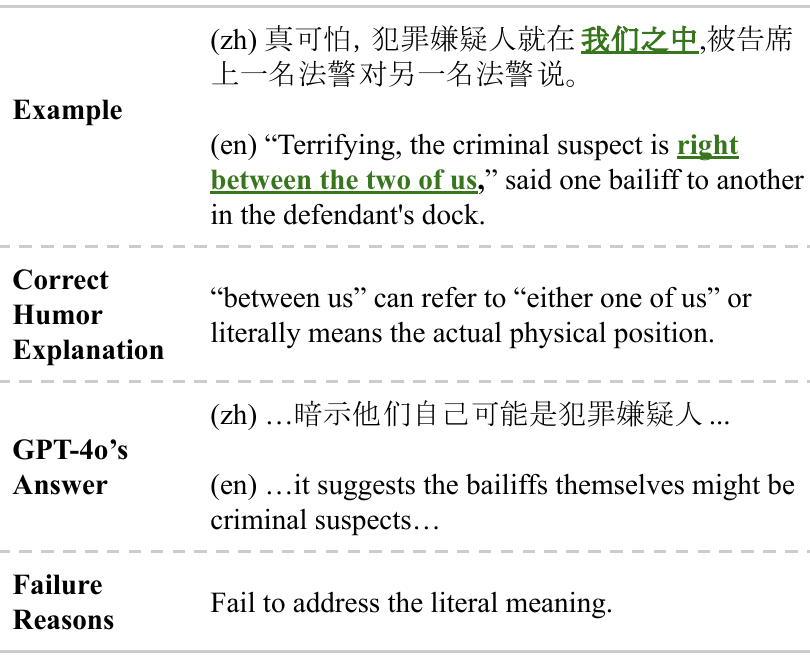}

\noindent LLMs may fail to ground their responses to the context when they explain the joke.
For instance, in this example, ``between us'' typically means ``either you or me'', but it also has the literal meaning to indicate the person standing ``between us'', which is the right interpretation given that the two bailiffs are talking about the criminal.
However, GPT-4o only reasons that ``the criminal is either you or me'' but fails to capture the literal meaning from the context.
We hypothesize that in the pre-training corpus, ``between us'' most likely acquires the meaning of ``either you or me'' rather than the literal meaning in a scenario like this, which biases the LLM to not reason the literal meaning needed for the explanation.


\paragraph{Hallucinations.}\leavevmode

\noindent\includegraphics[width=\linewidth]{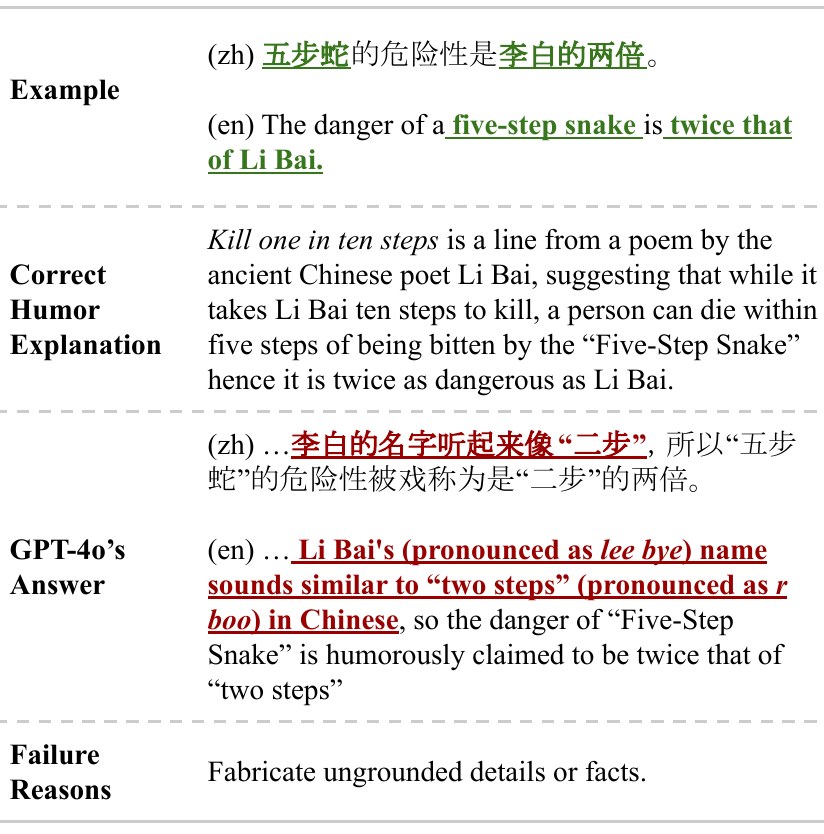}

\noindent LLMs may fabricate ungrounded details or facts in the joke explanation. 
For instance, in this explanation, GPT-4o provides the reason that ``Li Bai's name sounds similar to two steps'', while ``Li Bai'' (pronounced as {\it lee bye}) does not sound like ``two steps'' (pronounced as {\it r boo}).

On the other hand, the correct explanation requires an understanding of a Chinese poem from Li Bai, ``\chenv{十步杀一人}'' (The warrior kills a person for every ten steps). 
This sentence is intended to praise the courage of the soldiers, but the joke deliberately portrays this as a characteristic of Li Bai.
Therefore, compared to Li Bai who can kill a person in ten steps, a five-step snake, which can kill a person in five steps, is twice as dangerous as Li Bai.
Such explanation requires LLMs to have a deep understanding of Chinese culture and reason over these cultural terms, which poses a great challenge to the current LLMs.
Though recent works have made progress towards building LLMs beyond English \cite{du2024chinese, zhao2024llama}, building an LLM that can understand such nuanced Chinese cultural terms can be extremely hard.

\paragraph{Pun-based Humor.}\leavevmode

\noindent\includegraphics[width=\linewidth]{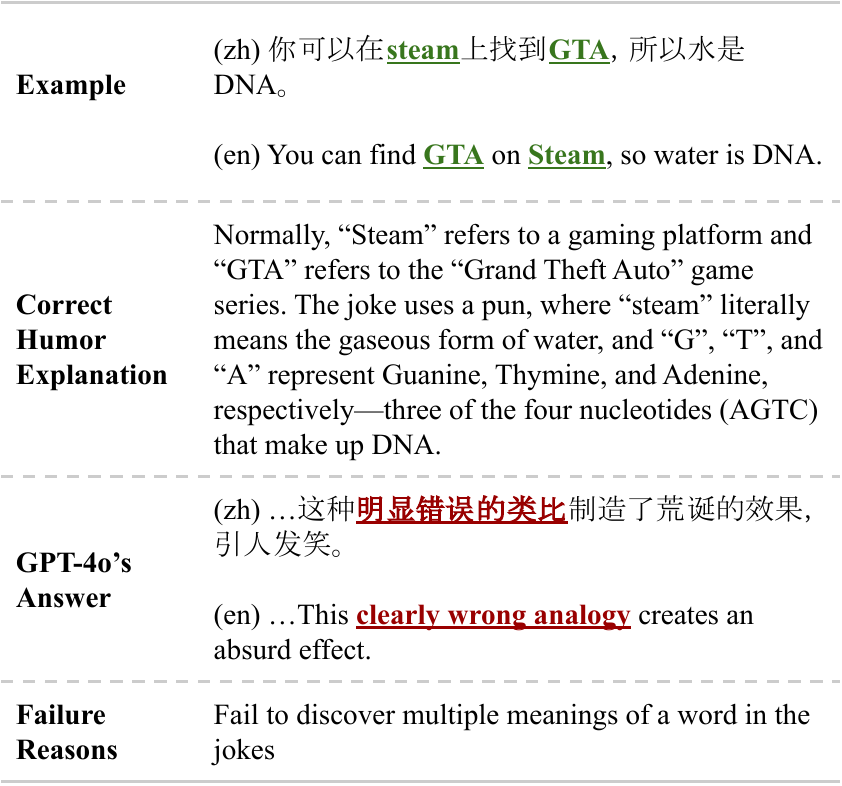}

\noindent LLMs may fail to discover multiple meanings of the same word in the joke, leading to its failure on pun-based jokes where the humor lies in inverting the conventional usage of words. 
In this example, GPT-4o fails to understand the switch from the video game ``Steam'', ``GTA'' to the scientific terminologies ``G'', ``T'', ``A'' that make up DNA. 
Typically, ``steam'' refers to a game platform, and ``GTA'' refers to the game series ``Grand Theft Auto''. 
The joke uses a pun on words as ``steam'' in its literal sense means water vapor, and ``GTA'' can represent not only the video game, but guanine, thymine, and adenine, which are nucleotides involved in the structure of DNA. 
These kinds of jokes require LLMs to identify puns, and reason on the association of the multiple meanings.
In addition, such a process requires LLMs to bridge the logic gap between these terms, such as ``steam'' and ``GTA'', and an unusual context, like ``water is DNA''. 
The overall process is particularly challenging, involving scientific knowledge and creative thinking which LLMs still struggle with.

\paragraph{Parsing Error.}\leavevmode

\noindent\includegraphics[width=\linewidth]{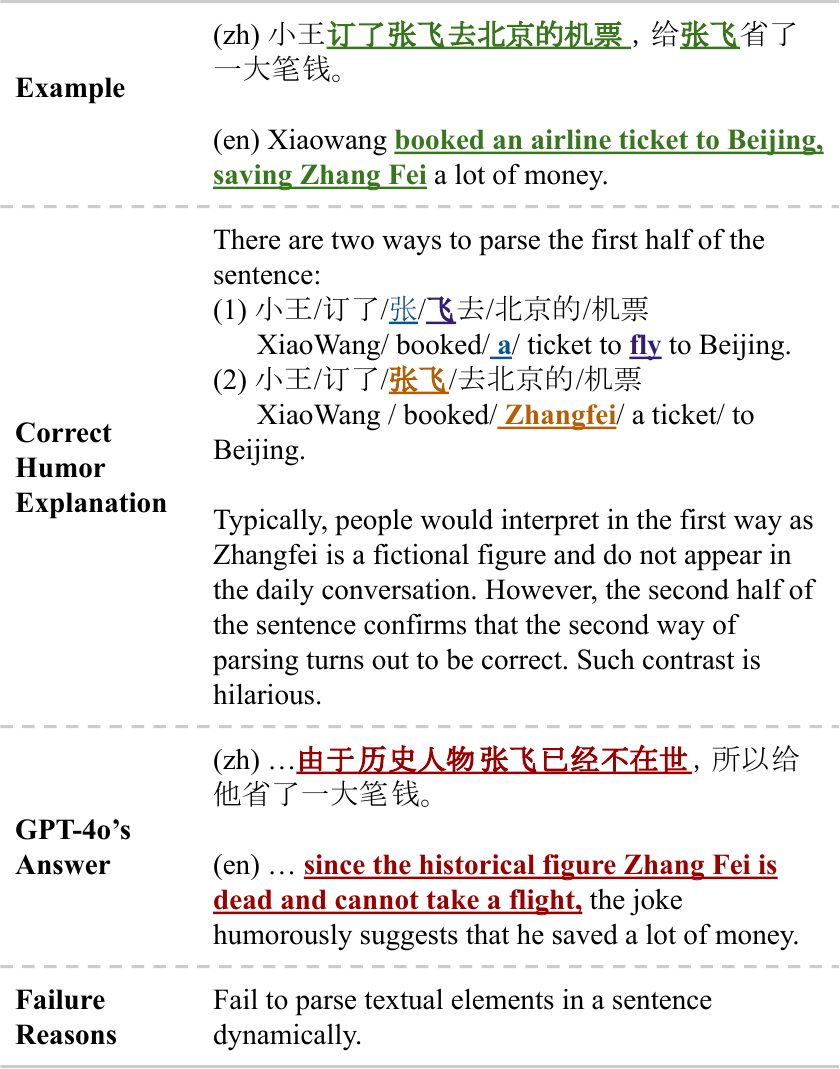}
LLMs cannot parse certain elements in the sentence in more than one way at a time, therefore failing to explain jokes that require different parsing for the same sentence. 
In this example, the humor hinges on the ambiguity of the phrase ``\chenv{张飞}'', which can be interpreted either as part of the verb phrase implying ``a ticket flying to Beijing'' or as a proper noun, referring to the historical figure Zhang Fei. 
There is great flexibility in how Chinese characters form a word, as each Chinese character can serve independently as a word, and combinations of these characters can also forge new words or phrases.
There are decades of research studying the problem of parsing Chinese \cite{sun-jurafsky-2004-shallow, sun-etal-2009-chinese}.
Recently, researchers have proposed task-specific tokenization approaches that adapt the parsing process to better align with downstream tasks \cite{liu-etal-2021-bridging, liu-etal-2023-task}.
However, how to incorporate different ways of parsing at one time still remains challenging.

\paragraph{Cross-lingual Humor.}\leavevmode

\noindent\includegraphics[width=\linewidth]{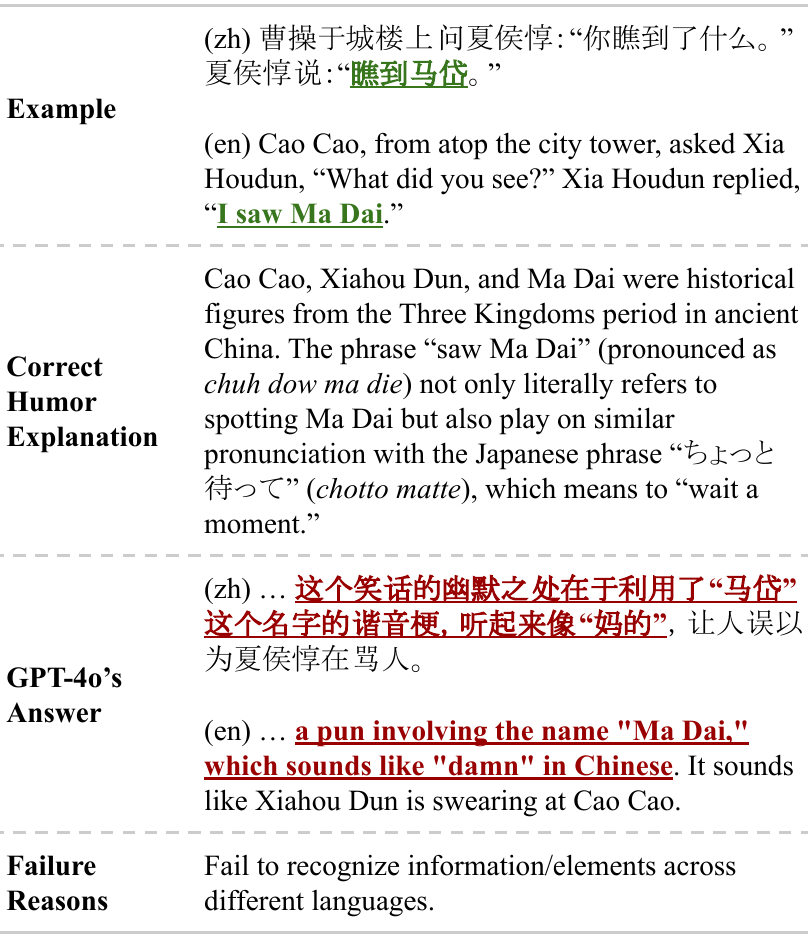}

\noindent LLMs may fail to recognize elements or information across different languages.
In its explanation, GPT-4o tries to link the pronunciation of ``Ma Dai'' to other Chinese terms but fails to reason on the similar pronunciations across the Chinese term ``\chenv{瞧到马岱}'' (pronounced as {\it chuh dow ma die}, meaning ``saw Ma Dai'')  and the Japanese term ``\raisebox{-0.15cm}{\includegraphics[width=0.3\linewidth]{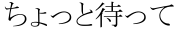}}'' ({\it chotto matte}, meaning ``wait a moment'').
Such examples require LLMs to connect pronunciations across languages, which may be rare in the LLMs' pre-training corpus and poses significant challenges to current LLMs.

\paragraph{Excessive Sensitivity.} 

For ERNIE Bot, apart from making mistakes in all the aforementioned categories, it also shows excessive sensitivity to certain examples.
Specifically, for content that contains languages related to hate speech but used in non-harmful contexts, ERNIE Bot refuses to answer.
During our evaluation, we observe excessive sensitivity in the ERNIE Bot's responses to humor related to medical ethics, and political discussions.
This suggests that how to correctly understand the context and the language toxicity remains an open challenge \cite{zhang2024dont}.
Such a problem is especially important for humor explanation, as misclassifying non-toxic context can lead to responses that deviate from the intended humor.


\section{Can LLMs Serve as Preference Annotator?}
\label{app-sec: llms-as-preference-annotator}

\begin{figure}
    \centering
    \includegraphics[width=\linewidth]{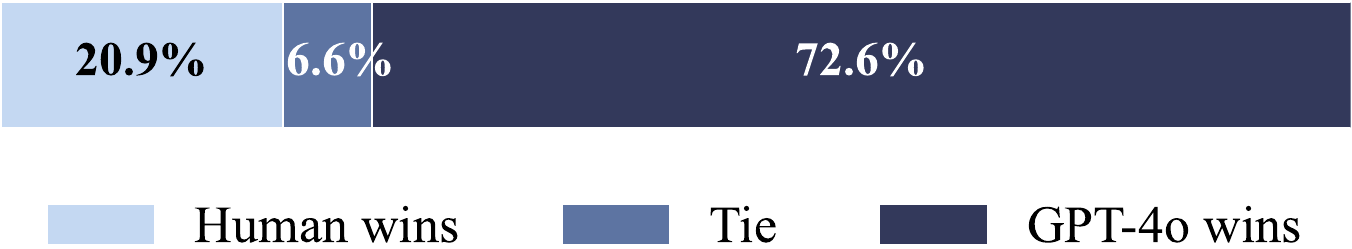}
    \caption{Preference annotation from GPT-4o.
    We prompt GPT-4o to choose a better explanation between its own explanation and the explanation written by human.
    We note that the GPT-4o's preference is significantly different from the human preference in \Cref{fig:gpt-4o-preference-eval}.}
    \label{fig:gpt-4o-annotator-eval}
\end{figure}

\Cref{fig:gpt-4o-annotator-eval} presents the results of GPT-4o preference study.
We prompt the GPT-4o by:

\chenv{\noindent``对于笑话``[Joke]''

\noindent以下两个解释

\noindent解释1: [Explanation 1]

\noindent解释2: [Explanation 2]

\noindent哪个解释更好的解释了这个笑话的幽默之处？}'',

\noindent which translates to,

\noindent ``For the joke``[Joke]''

\noindent The following two explanations

\noindent Explanation 1: [Explanation 1]

\noindent Explanation 2: [Explanation 2]

\noindent Which one better explains the reason for why this joke is funny?''

Though prior works have shown the promise of employing LLMs as evaluators \cite{fu2023gptscore, liu-etal-2023-g}.
We note that as the evaluator for humor explanation, GPT-4o's preference significantly differs from the human preference compared to \Cref{fig:gpt-4o-preference-eval}. 
Our results show that GPT-4o significantly favors its own explanation than explanation from human, which aligns with the finding in previous study that LLM evaluators prefer their own answer \cite{panickssery2024llm}.
But for a small proportion of the dataset, GPT-4o prefers explanation from human than the explanation from itself, and \Cref{fig:example} demonstrates one such example, which shows the potential of leveraging LLMs in automatic humor evaluation.

\section{Potential Application}
\label{app-sec: potential-application}
First, we expect researchers to use \dataname~to comprehensively evaluate the humor understanding abilities of LLMs.
We want to stress that this may not be an effort limited to the NLP community, but researchers from psychology, cognitive science and beyond can also use \dataname~to study how human perceives humor differently than LLMs.
Such insights can in turn guide the algorithm innovation in the NLP commuinty on humor understanding.

In addition, \dataname~contains comprehensive linguistic and cultural components.
We expect researchers to use \dataname~to evaluate the culture-understanding abilities of current LLMs, and develop LLMs that can deeply understand the nuances of diverse cultural backgrounds.

We also consider \dataname~to be especially valuable to enhance the logic reasoning abilities of LLMs.
As revealed by \citet{bai2024coig}, RZB data is effective in improving LLMs' ability on Chinese reasoning that is not limited to humor understanding, but can be generalized to many other tasks.

Last but not least, we hope that \dataname~facilitates humor evaluation, especially humor evaluation in non-English languages like Chinese.
As discussed in \Cref{app-sec: llms-as-preference-annotator}, our evaluation reveals a significant gap between LLMs' preference versus human preference, we expect researchers to come up with new algorithms to better evaluate humor reasoning automatically.

\end{document}